\documentclass[letterpaper, 10 pt, conference]{ieeeconf}  
\usepackage{subfigure}

\IEEEoverridecommandlockouts                              

\usepackage{booktabs}                   
\usepackage{graphicx}      
\usepackage[ruled,vlined]{algorithm2e}          
\usepackage{color}
\usepackage{algorithmic,algorithm2e,float}
\usepackage{multirow}

\title{\LARGE \bf
RADA: Robust Adversarial Data Augmentation for Camera Localization in Challenging Weather}

\author{Jialu Wang$^{1}$, Muhamad Risqi U. Saputra$^{2}$, Chris Xiaoxuan Lu$^{3}$, Niki Trigoni$^{1}$, and Andrew Markham$^{1}$ 
\thanks{$^{1}$Jialu Wang, Niki Trigoni, and Andrew Markham are with the Department of Computer Science, University of Oxford, UK {\tt\small \{jialu.wang, niki.trigoni, andrew.markham\}@cs.ox.ac.uk}}
\thanks{$^{2}$Muhamad Risqi U. Saputra is with the Data Science Department at Monash University Indonesia {\tt\small risqi.saputra@monash.edu}}
\thanks{$^{3}$Chris Xiaoxuan Lu is with the School of Informatics at the University of Edinburgh, UK {\tt\small xiaoxuan.lu@ed.ac.uk}}
}

\begin{document}

\maketitle
\thispagestyle{empty}
\pagestyle{empty}
\begin{abstract}

Camera localization is a fundamental and crucial problem for many robotic applications. In recent years, using deep-learning for camera-based localization has become a popular research direction. However, they lack robustness to large domain shifts, which can be caused by seasonal or illumination changes between training and testing data sets. Data augmentation is an attractive approach to tackle this problem, as it does not require additional data to be provided. However, existing augmentation methods blindly perturb all pixels and therefore cannot achieve satisfactory performance. To overcome this issue, we proposed RADA, a system whose aim is to concentrate on perturbing the geometrically informative parts of the image. As a result, it learns to generate minimal image perturbations that are still capable of perplexing the network. We show that when these examples are utilized as augmentation, it greatly improves robustness. We show that our method outperforms previous augmentation techniques and achieves up to two times higher accuracy than the SOTA localization models (e.g., AtLoc and MapNet) when tested on `unseen' challenging weather conditions. \\


\end{abstract}

\section{Introduction}

\begin{figure*}[htbp]
	\centering
    \includegraphics[width=0.9\textwidth]{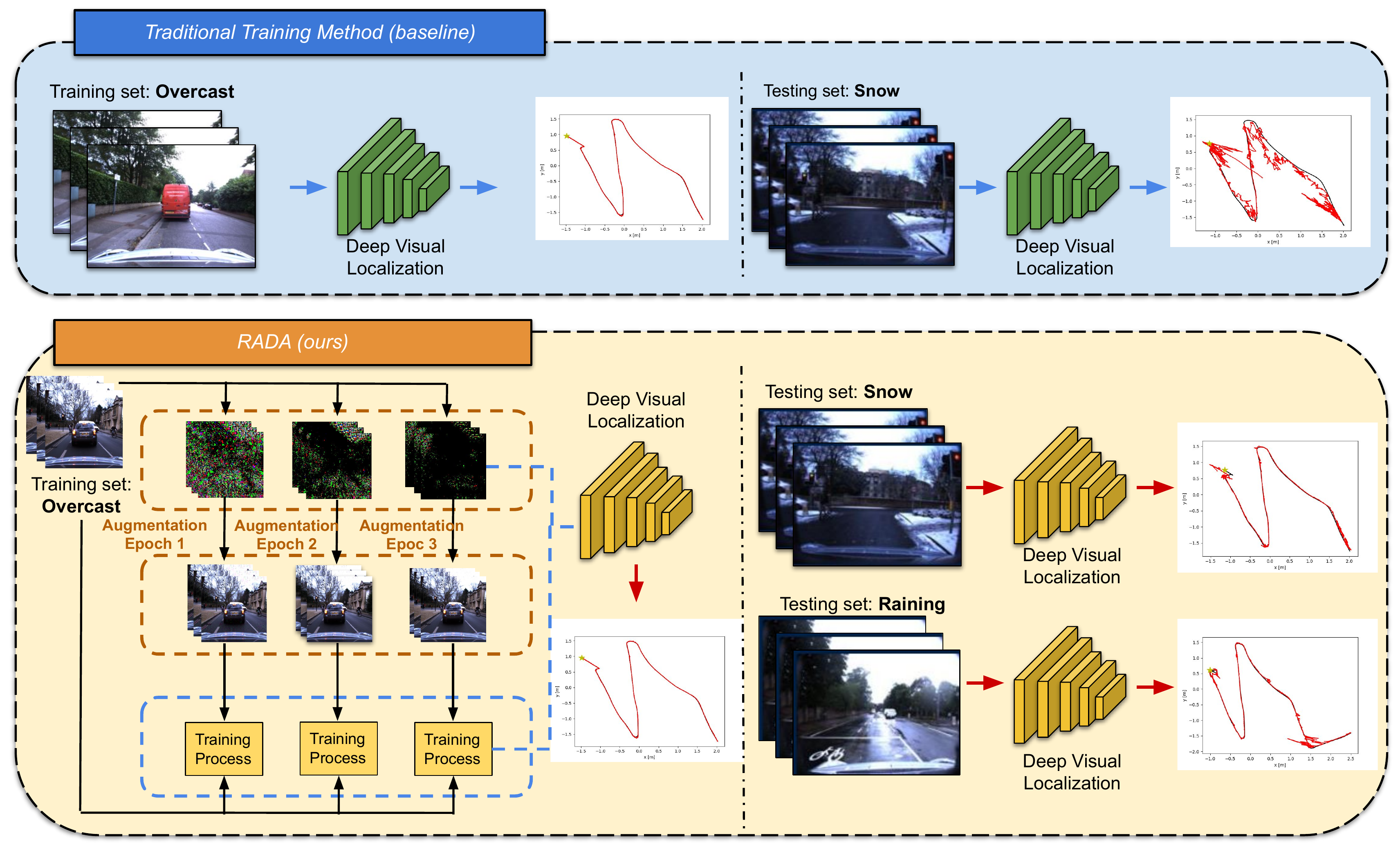}
	\caption{ \textbf{Upper}: Traditional training method for deep camera localization models. \textbf{Lower}: Our proposed training method with RADA.}
\label{fig:RADA_working_mechanism}
\end{figure*}

Camera localization refers to the problem of recovering the 3D position and orientation of a camera given an image or a set of images as the input. This problem is fundamental and crucial for many robotic applications such as autonomous driving, domestic robot navigation, augmented reality, and so on. Camera localization system was initially developed based on hand-crafted features. However, the extracted features were vulnerable to cross-weather, such as lighting and saturation \cite{brachmann2017dsac}. Recently, deep-learning based models were proposed to alleviate those problems by automatically extracting informative features from images \cite{brahmbhatt2018geometry, hong2020radarslam, wang2020atloc}. However, they remain less robust towards real-world scene variations, especially when the model is tested on unseen scenarios with a domain shift, such as a change in ambient weather, impacting appearance and illumination \cite{shuadversarial}.

One possible solution to improve the robustness of existing deep learning-based localization models is to use basic data augmentation to diversify the training data distribution. For example, \cite{brahmbhatt2018geometry, wang2020atloc} shifted the RGB pixel values, \cite{maddern2020real} used Gaussian noise, and \cite{sobh2021adversarial} employed blur, HSV shift, and other perturbations to simulate cross weather interference. However, these domain adaptation techniques are \textit{data agnostic}. They perturb all image pixels in a uniform fashion, corrupting important information for camera localization.

In recent studies \cite{ren2020adversarial, xie2019feature, carlini2018ground}, adversarial training is considered as one of the most promising ways to improve robustness especially for classification problems \cite{madry2017towards, kannan2018adversarial}, \cite{croce2020adversarial}, \cite{zunair2021synthesis, hussein2020augmenting}. Nevertheless, it has not been used in camera localization under domain shift. \cite{shuadversarial} used standard data augmentation methods (e.g. Gaussian blur) to improve the robustness of steering angle predictions but then they utilize `adversarial training' to fine-tune the model's hyper-parameters (e.g., variance of Gaussian). Recently, \cite{sobh2021adversarial} evaluated the performance of various adversarial attacks on autonomous driving data, demonstrating the brittleness of many approaches. Inspired by this paper, we take a step further, and  explore if these attacks can be used in camera localization so as to achieve greater robustness for cross-weather adaptation. Nevertheless, our experiments in Section \ref{Overall_Performance} indicates that directly adopting the existing adversarial training methods blindly perturb all pixels and yield unsatisfactory performance. 

In this paper, we propose a novel adversarial data augmentation system for robust camera localization, dubbed as \textbf{RADA}. Instead of corrupting the entire image uniformly, RADA concentrates on perturbing only the geometrically informative parts of the image, especially those extensively used by camera localization such as trees, building landmarks, etc. In that sense, RADA generates perturbation that can greatly reduce the performance of localization models while at the same time minimally altering the original image. We demonstrate how this seemingly paradoxical minimal perturbation actually yields more robust and noise-resistant performance in reality when faced with cross-domain interference and unseen weather conditions. In summary, our key contributions are given as follows:
\begin{itemize}
    \item We propose RADA, a \textbf{R}obust \textbf{A}dversarial \textbf{D}ata \textbf{A}ugmentation system for camera localization. RADA focuses on perturbing the geometrically informative part of images such that it can significantly improve the robustness of the localization models. To the best of our knowledge, this is the first work on adversarial learning for camera localization. 
    \item We demonstrate the effectiveness of RADA by applying it on two state-of-the-art localization models, namely AtLoc \cite{wang2020atloc} and MapNet \cite{hong2020radarslam}, such that their accuracy can be improved up to more than two fold, especially when they were tested on unseen weathers.
    \item We perform extensive experiments on the public Oxford RobotCar dataset, including analysing the trade-off between the quality of the training data and the incremental gain of RADA.
\end{itemize}

\section{Related Work}
\subsection{Data Augmentation Approaches}
Data augmentation is a technique to increase the scale and the complexity of training dataset in order to improve the model robustness \cite{cygert2020toward}. This is achieved by introducing artificially generated data samples without changing their corresponding labels, and without requiring data from the target domain to be provided. \cite{shorten2019survey} divides the existing computational visual data augmentation techniques into two categories: basic image manipulation and deep-learning approaches.

\paragraph{Basic image manipulation}
These are a group of approaches that either expand the data set in a relatively random and blind manners, such as by employing color space or geometric transformations, or give a strong assumption on the noise distribution such as by using Gaussian noise. S. Brahmbhatt et al. \cite{brahmbhatt2018geometry} simulated the interference caused by the change of time and weather in their MapNet paper by adjusting the brightness of images (via ColorJitter toolbox). Compared to learning-based augmentation approaches, improvement in system robustness is limited.

\paragraph{Deep-learning approaches}
Generative Adversarial Networks (GAN) and adversarial training are two different deep-learning based data-augmentation approaches. GAN approaches \cite{goodfellow2014generative} utilize a generator network to attempt to fool the classifier in the training phase. The original dataset is then artificially augmented by the generator via the learned data distribution. \cite{madry2017towards} showed the two methods are closely related and can even be combined to achieve complementary advantages. 

\subsection{Adversarial Training}
Adversarial training is a deep-learning based data augmentation technique that directly augments the data by maximizing the loss function of the classifier \cite{lowd2005adversarial, goodfellow2018making}. It was originally proposed by \cite{goodfellow2014explaining} based on the principle of gradient ascent. They multiply the sign of the gradient by a very small step $\varepsilon$ to obtain the perturbation, and named this attack FGSM. \cite{kurakin2016adversarial} proposed FGM which replaces the sign of the gradient with the unit vector of gradient direction to avoid the fact that different pixels are perturbed by the same step size. PGD \cite{madry2017towards} optimized the perturbation mechanism with an extra constraint through multiple iterations. Subsequently, EAT \cite{tramer2017ensemble} and ALP \cite{kannan2018adversarial} improved training efficiency by increasing sample diversity and using `pairing loss' respectively.

\begin{figure}[htbp]
	\centering
    \includegraphics[width=0.48\textwidth]{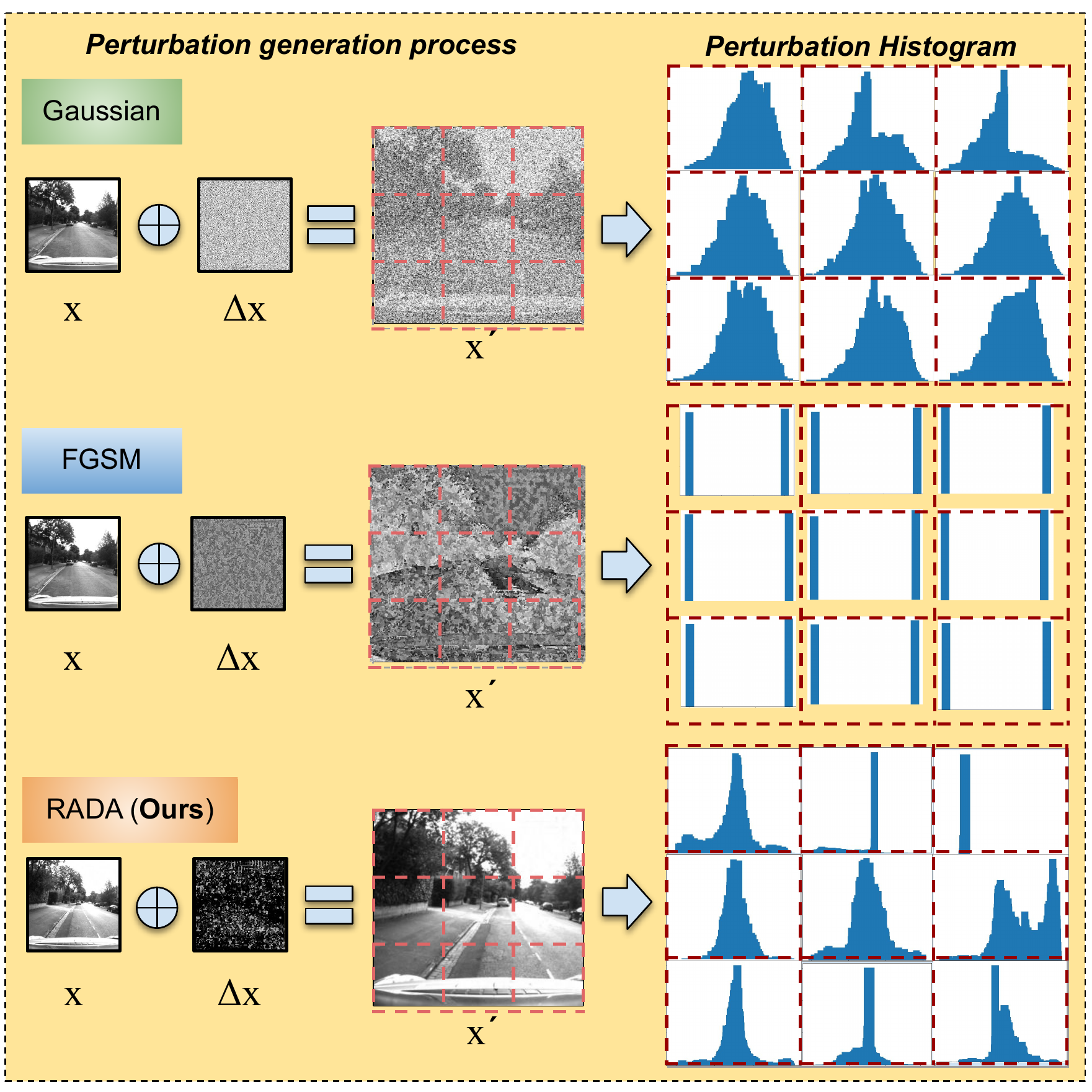}
	\caption{The comparison of perturbation results between Gaussian, FGSM \cite{kurakin2016adversarial}, and RADA (ours). $x$, $\Delta x$, and $x'$ are the original image, the generated perturbations, and the perturbed image respectively. The histogram of perturbations are obtained by equally dividing $\Delta x'$ into 9 sub-squares and computing the frequency (note that pixels without interference are omitted. The Gaussian perturbed outputs were clipped within [0,1]. ). Both Gaussian and FGSM histograms indicate that their perturbations are distributed uniformly in all regions. In contrast, the perturbations generated by RADA are  concentrated in the regions which contain discriminative geometrical structures that are important for localization (e.g., trees, buildings). By training the localization model on $x'$, the network does not overfit to weather-specific structure and thus the robustness to cross weather variations will be significantly improved.}
\label{fig:histogram}
\end{figure}

\section{Methods}
\subsection{Background}
The mathematical expressions for generating the (augmented) adversarial samples is illustrated as follows. Let $\theta$ be the model $g_\theta$'s weights, $L$ is the loss function between the labels $y$ and the predictions $g_\theta(x)$, $\widetilde x$ is the generated adversarial sample, and the small constant $\varepsilon$ describes the step size of the perturbation. The problem of generating adversarial samples boils down to finding a set of perturbed input images $\widetilde x$, which makes the loss function $L$ as large as possible as follows:
\begin{equation}
    \widetilde x = \mathop{argmax}\;L(g_\theta(x), y) \label{con:1}
\end{equation}
Eq. (\ref{con:1}) can be simplified by solving the following first order Taylor expansion:
\begin{equation}
    \widetilde x =\mathop{argmax}\limits_{\widetilde x: ||\widetilde x - x||_{norm}\leq\varepsilon}{(\bigtriangledown_x L(g_\theta (x), y))^T(\widetilde x-x)} \label{con:2}
\end{equation}
When $norm = \infty$, the perturbation $r_{adv}$ and the adversarial sample $\widetilde x$ then can be obtained by:
\begin{equation}
    \widetilde x = x + r_{adv} = x + \varepsilon\;\mathop{sign}(\bigtriangledown_x L(g_\theta (x), y)) \label{con:3}
\end{equation}

\begin{algorithm} [h]
\caption{RADA Algorithm}
\algsetup{linenosize=\tiny}
	\SetAlgoLined
\For{each epoch}{
		1. Use the range of $x$ to calculate the perturbation threshold $\eta_{th}$ (Eq. (\ref{con:5})) \;
		2. Freeze the model parameter $\theta$, obtain $L_{\theta}(p, p^*)$ (Eq. (\ref{con:6}) and (\ref{con:7})), and calculate the gradient $g$ (Eq. (\ref{con:8}))\;
		3. Calculate $r_{adv}$ by using $\mathop{g}$ (Eq. (\ref{con:9})). If any perturbation value exceeds the threshold $\eta_{th}$, assign $\eta_{th}$ to that perturbation\;
		4. Calculate adversarial sample $x_{adv}$ using $r_{adv}$ (Eq. (\ref{con:10}))\;
		5.  If any pixel value in $x_{adv}$ exceeds the range $[0,255]$, assign the nearest boundary (0 or 255) to that pixel\;
		6.  Use $x_{adv}$ to train the model and update the weights $\theta$\;
		}
\end{algorithm}

\subsection{Proposed Perturbation Mechanism}
The perturbation $r_{adv}$ in Eq. (\ref{con:3}) can be redefined into Eq. (\ref{con:4}) where $pow = 0$ as follows:
\begin{equation}
    r_{adv} = \varepsilon \times sign(g) \times g^{pow}, \qquad pow = 0 \label{con:4}
\end{equation}
As described, existing methods either take $pow = 0$ (e.g., FGSM) or $pow = 1$ (e.g., FGM, PGD, etc.). However, our experiments show that directly using this traditional adversarial training approach cannot satisfactorily improve the robustness (see Sec. \ref{Overall_Performance}). This is because all pixels will receive the maximum perturbation along the gradient direction which corrupts all pixel values, whether they are key for localization or not, generating a scattering of confounding dots (see Fig. \ref{fig:histogram}). On the contrary, our aim is to concentrate on perturbing only the geometrically informative parts of the image. As a result, the method should be able to learn to generate minimal perturbation whilst greatly reducing the localization performance. Then, when the models are retrained using these perturbed images in addition to the original training set, the robustness will improve significantly. To this end, the following changes have been made to the traditional methods:

(1) \textbf{Adversarial perturbation}. 
In our approach, we take $pow \textgreater 1$ and a relatively larger $\varepsilon $ in Eq. (\ref{con:4}) which enables RADA to not only take the direction of the gradient, but to also learn how sensitive the network is to a particular pixel, and then focuses the perturbation efforts towards that pixel. As a result, it can concentrate on perturbing small geometrically informative regions of an image as seen in Fig. \ref{fig:histogram}.

(2) \textbf{Threshold: perturbation constraint mechanism}. Before each batch of training, we calculate a perturbation threshold from the pixel range so as to limit the size of the perturbation. This design enables the perturbation to be adaptive to the input dataset. 
As shown in Eq. (\ref{con:5}) below, $\eta_{th}$ is the threshold used to limit the perturbation and $x_{max}$ and $x_{min}$ are the maximum and minimum pixel values of each batch. The constant $\eta$ is the number of thresholds, which is positively correlated with the accuracy.
\begin{equation}
    \eta_{th} =(x_{max}-x_{min}){\div} \eta 
    \label{con:5}
\end{equation}

(3) \textbf{Clipping: Adversarial sample constraint mechanism}.
To avoid invalid values, we further clip the value of the perturbed pixels into the range of [0,255]. Fig.\ref{fig:ablation_study} (d) shows that this restriction can effectively reduce the confounding pixels.

(4) \textbf{Untargeted perturbation}. Unlike the targeted perturbation approaches \cite{shorten2019survey} (e.g., by turning a sunny image into a rainy image such that the method is robust to the rainy image) which typically fail to adapt to `unseen' conditions (e.g., snowy conditions), our approach uses untargeted perturbation  \cite{sobh2021adversarial}. Therefore, it can adapt to a wide variety of unseen weather conditions without specifying the target domain.
\subsection{Proposed Adversarial Training for Camera Localization}
A deep camera localization model is trained to predict the 6-DoF camera poses ${ p=(t,w)}$ by inputting an RGB image $x$ and learning the model weights $\theta$ through back-propagation algorithm. Given the estimated camera pose $p$ and its corresponding ground truth $p^*$, the loss function $L_{\theta}(p, p^*)$ is defined as follows:
\begin{equation}
    L_{\theta}(p, p^*) = \sum_{i=1}^{|D|}h(p_i, p_i^*)+\alpha \sum_{i,j=1,i\not=j}^{|D|}h(v_{i,j}, v_{i,j}^*)
     \label{con:6}
\end{equation}
\begin{equation}
    h(p, p^*)=||t-t^*||_1e^{-\beta}+\beta+||w-w^*||_1e^{-\gamma}+\gamma \label{con:7}
\end{equation}
where $i$,$j$ refers to the indexes of two adjacent images, $v_{i,j} = (t_i-t_j,w_i-w_j)$ is the relative pose between $p_i$ and $p_j$, and $h\{.\}$ is the difference between $p$ and $p^*$. $\beta$ and $\gamma$ are trainable parameters used to balance the loss between translation $t$ and rotation $w$. Let $L_{\theta}^r(p, p^*)$ be the loss function in Eq. (\ref{con:6}) after freezing the weights $\theta$, then the gradient $g$ is defined as the derivative w.r.t. $x$ as follows:
\begin{equation}
    g = \bigtriangledown_x L_{\theta}^r(p, p^*) \label{con:8}
\end{equation}
Based on the gradient ascent principle \cite{goodfellow2014explaining}, the perturbation $r_{adv}$ is equal to the gradient direction multiplied by a constant step $\varepsilon$. By plugging in Eq. (\ref{con:8}) into Eq. (\ref{con:4}), then we obtain our perturbation for camera localization as follows:
\begin{equation}
    r_{adv} = \varepsilon \times sign(\bigtriangledown_x L_{\theta}^r(p, p^*)) \times (\bigtriangledown_x L_{\theta}^r(p, p^*))^{pow}, pow>1\label{con:9}
\end{equation}
The generated adversarial sample $x_{adv}$ can be defined as:
\begin{equation}
    x_{adv} = x + r_{adv} \label{con:10}
\end{equation}
Finally, the overall workflow of generating RADA adversarial samples is shown in Algorithm 1. The RADA system pipeline is illustrated in Fig. \ref{fig:RADA_working_mechanism}.





\begin{table}[h]
\centering
\caption{RobotCar Dataset Selection}\label{tab:Dataset_sequence}
\begin{tabular}{cccc}
\toprule 
Sequence& Time&	Tag& Model\\
\toprule   
loop& 2014-06-26-09-24-58& 	overcast& Training\\
loop& 2014-06-26-08-53-56& 	overcast& Training\\
loop& 2014-06-23-15-36-04& sunny& Testing\\
loop& 2014-06-24-14-09-07& 	over-exposure& Testing\\
\hline
fullA& 2014-11-28-12-07-13& 	overcast& Training\\
fullA& 2014-12-02-15-30-08& 	overcast& Training\\
fullA& 2014-11-25-09-18-32& 	rain& Testing\\
\hline
fullB& 2015-02-13-09-16-26& 	overcast& Training\\
fullB& 2015-02-03-08-45-10& 	snow& Testing\\
\bottomrule 
\end{tabular}
\end{table}

\section{EXPERIMENTS}
\subsection{Localization model and Evaluation metrics}
In order to demonstrate the efficacy of our adversarial data augmentation, we use RADA to enhance two state-of-the-art camera localization models, namely MapNet \cite{hong2020radarslam} and AtLoc \cite{wang2020atloc}.
\paragraph{MapNet}
MapNet is a DNN for localization. MapNet is a recent deep learning based camera localization model that can automatically extract features and directly recover the absolute camera pose from input images. It achieves a relative higher robustness to illumination changes with an additional geometric constraint  between  pairs  of  frames. 

\paragraph{Atloc} AtLoc \cite{wang2020atloc} is a SOTA camera localization model, which shows high robustness to dynamic objects and changing illumination by introducing a self-attention mechanism. 

For a fair comparison, in this paper, we used the same evaluation strategy for both MapNet and AtLoc, i.e. mean translation error (m) and mean rotation error (degree).

\subsection{Dataset}
Oxford RobotCar dataset \cite{maddern20171} was collected from a 10km route in an autonomous Nissan LEAF car through central Oxford over a year. The collected data contains most of weather conditions, including rainy, snowy, over-exposure, and nighttime, which makes it particularly challenging for camera localization problem. Details about selected sequences can be found in Table. \ref{tab:Dataset_sequence}. For each sequence, we adopted input images recorded by the stereo centre camera sequence 01 with a resolution of 1280 x 960. The interpolations of INS data were labelled as the corresponding ground truth. 

\begin{table*}[h]
\centering
\caption{Comparing RADA with SOTA methods using MapNet and AtLoc}\label{tab:SOTA_exp}
\begin{tabular}{c|c|c|c|c|c|}
\toprule 
Testing Sequence& AtLoc \cite{wang2020atloc} & AtLoc with Gaussian& AtLoc with FGSM& AtLoc with RADA (\textbf{ours})\\
\toprule   
sunny& 8.86m, $4.67^{\circ}$& 10.93m, $5.97^{\circ}$& 8.11m, $3.60^{\circ}$& \textbf{6.93m}, $\mathbf{3.40^{\circ}}$ \\
\toprule
over-exposure& 22.17m, $17.72^{\circ}$& \textbf{8.80m}, $9.38^{\circ}$ & 9.54m, $8.76^{\circ}$ &9.23m, $\textbf{8.84}^{\circ}$\\
\toprule
rain& 8.99m, $2.15^{\circ}$&12.40m, $2.96^{\circ}$ & 12.90m, $2.74^{\circ}$ & \textbf{6.87m, $\textbf{1.90}^{\circ}$} \\
\toprule
snow& 37.99m, $8.18^{\circ}$& 29.21m, $10.82^{\circ}$ & 24.56m, $11.12^{\circ}$ &\textbf{13.07m}, $\textbf{8.15}^{\circ}$\\
\toprule 
average& 19.50m, $8.18^{\circ}$& 15.34m, $7.28^{\circ}$& 13.78m, $6.55^{\circ}$& \textbf{9.02m}, $\textbf{5.57}^{\circ}$\\
\bottomrule 
\toprule 
Testing Sequence& MapNet \cite{hong2020radarslam} & MapNet with Gaussian& MapNet with FGSM& MapNet with RADA (\textbf{ours})\\
\toprule   
sunny& \textbf{9.84m}, $\textbf{3.96}^{\circ}$& 69.48m, $50.22^{\circ}$& 16.89m, $11.65^{\circ}$& 17.26m, $6.82^{\circ}$\\
\toprule
over-exposure&  18.49m, $12.88^{\circ}$& 49.80m, $39.33^{\circ}$ & 19.74m,  \textbf{8.27}$^{\circ}$&\textbf{16.89m}, $11.65^{\circ}$\\
\toprule
rain& 15.00m, $3.72^{\circ}$& 16.11m, $3.60^{\circ}$ & 13.39m, $3.60^{\circ}$&\textbf{12.10m}, \textbf{2.56}$^{\circ}$\\
\toprule
snow&29.16m, $9.06^{\circ}$& 36.22m, $12.04^{\circ}$ &23.18m, $8.63^{\circ}$ &\textbf{12.91m}, $\textbf{2.65}^{\circ}$\\
\toprule 
average&18.12m, $7.41^{\circ}$&42.90m, $26.30^{\circ}$&18.3m, $8.04^{\circ}$&\textbf{14.79m}, $\textbf{5.92}^{\circ}$\\
\bottomrule 
\end{tabular}
\end{table*}

\begin{figure*}[htbp]
	\centering
\includegraphics[width=1\textwidth]{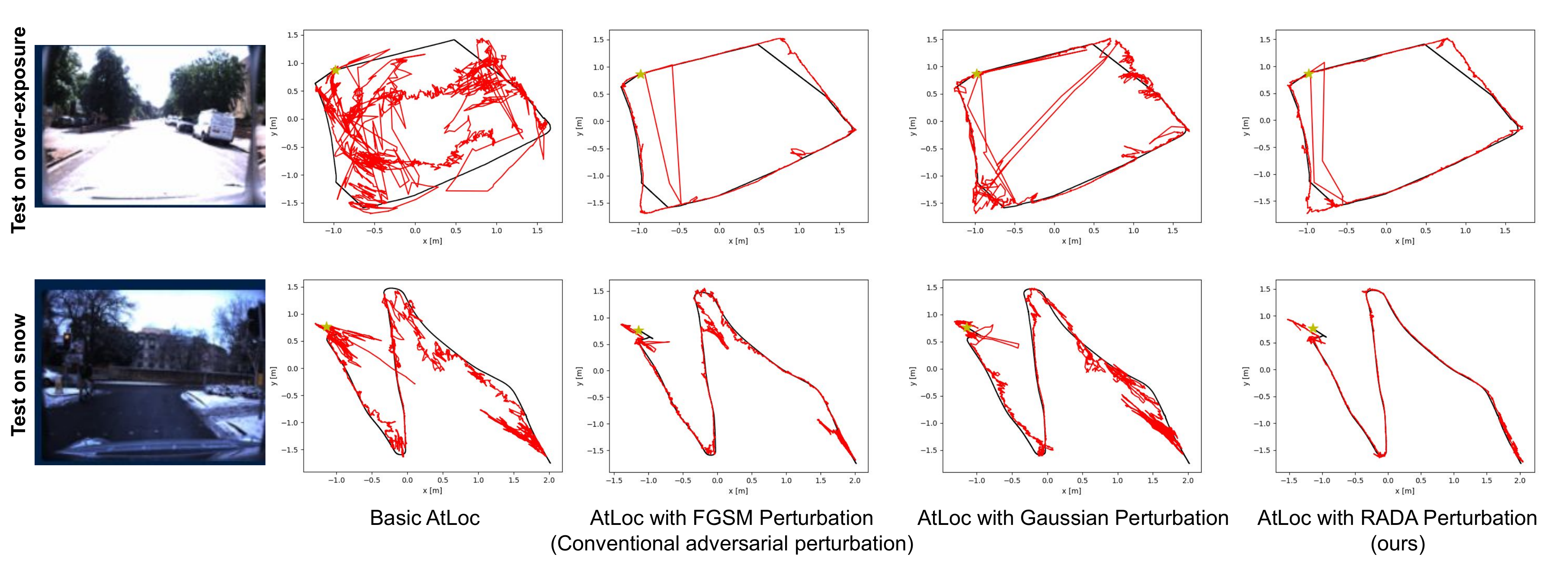}
	\caption{Trajectories on rainy day (upper) and snowy day (lower) test set.The ground truth (black line), predicted trajectories (red line) and starting point (start) are visualized above.}
	\label{fig:Trajectories}
\end{figure*}

\subsection{Implementation Details}
For all RADA adversarial perturbations, we set the power of the gradient $pow = 1.5$, the step size of the perturbation $\varepsilon$ = 158 (Eq. (\ref{con:9})), and the number of threshold $\eta$ = 10 (Eq. (\ref{con:5})). For all the FGSM perturbations, we set $\varepsilon$ = 0.3, which is a typical value for this method. For the Gaussian perturbation, we set a commonly used distribution whose mean and variance are 0 and 0.05 respectively. All models were trained until convergence. For a fair evaluation, all challenging test weathers are hidden/excluded from the training data.

\subsection{Overall Performance}
\label{Overall_Performance}
 For all evaluations, we trained both AtLoc and MapNet on good weather dataset (overcast) by using different data augmentation methods, and then tested them in good weather (sunny) as well as 'unseen' challenging weathers (over-exposure, rainy, and snowy days). For non-augmented system, we trained basic AtLoc and MapNet by utilizing the original, unmodified dataset. For naive augmentation system, we add Gaussian noise to the training dataset. For conventional deep learning-based augmentation system, we employ FGSM \cite{goodfellow2014explaining} to perturb the dataset. Apart from the basic AtLoc and MapNet, all models were trained by using both the original and the perturbed data. Fig. \ref{fig:Trajectories} visualizes some of the predicted trajectory for AtLoc. 
 
 Table \ref{tab:SOTA_exp} shows the numerical evaluation results in terms of mean translation error (m) and mean rotation error (degree). All competing systems work properly when they were tested on the good weather data (sunny), but they suffer from accuracy degradation on other challenging conditions. Unmodified AtLoc and MapNet degrade significantly when they were tested on rainy, snowy and over-exposure data. Sometimes, Gaussian and FGSM can improve the network's robustness, but most of the time they confuse the network since the generated perturbation is either random or too noisy such that it corrupts critical pixel information used for camera localization. In contrast, AtLoc and MapNet modified with RADA outperform the other techniques on most of good and challenging weather conditions. This owes to its ability to focus the perturbation on geometrically important features for localization such that 1) the perturbed image resembles the original image, keeping important information for localization, and 2) the perturbation forces the network to not overfit to any weather-specific structures, increasing the robustness to cross weather variations as described in Fig. \ref{fig:histogram}. On average, RADA yields $36.7\%$ and $26.5\%$ smaller translation and rotation error respectively for both AtLoc and MapNet.
 
\begin{table*}[h]
 \centering
 \caption{Analysis on the Incremental Gain of RADA}\label{tab:partial_rain}
  \begin{tabular}{c|c|c|c|c}
  \hline
  
  \multirow{2}{*}{{\% rainy data in training set}} & \multicolumn{2}{|c|}{Test on rainy data} &\multicolumn{2}{|c}{Test on sunny data}\\\cline{2-5}
  &base AtLoc&RADA&base AtLoc&RADA\\\hline
  0\%&38.04m, $46.6^{\circ}$&34.57m, $55.17^{\circ}$&3.93m,  $7.14^{\circ}$&5.88m, $8.12^{\circ}$\\\hline
  20\%&28.28m, $24.44^{\circ}$&16.29m, $5.29^{\circ}$&17.67m, $10.61^{\circ}$&10.74m, $4.61^{\circ}$\\\hline
  40\%&24.1m,  $29.87^{\circ}$&10.97m, $7.55^{\circ}$&21.07m,,$40.36^{\circ}$&13.78m, $31.47^{\circ}$\\\hline
 \end{tabular}
\end{table*}   
  
 \subsection{Perturbation Histogram}
 To compare the perturbation results from different methods, the perturbation histograms are plotted as seen in Fig. \ref{fig:histogram}. As shown in Fig. \ref{fig:histogram}, the perturbation of Gaussian is normally distributed on the whole image, although it is heavily clipped in the brighter areas of the image. FGSM perturbation follows binary distribution in all regions as it perturbs every pixel with the same step size regardless the gradient scale. In general, both FGSM and Gaussian perturbations are evenly distributed on the whole image, generating a lot of noises and confounding pixels. On the contrary, RADA acts in a very different manner. As shown in Fig. \ref{fig:histogram}, RADA produces very wide and large perturbations over trees and buildings, but relatively less perturbations over sky and road. This is because RADA does not only take the direction of the gradient, but also learns how sensitive the network is to a particular pixel and then it focuses the perturbation efforts towards that pixel. In particular, RADA targets its perturbation on small geometrically informative regions rather than polluting the whole image. In that sense, RADA learns that road itself or the sky, which exists in most of the scenes, are not spatially discriminative features for camera localization. However other features like trees and buildings are important to estimate camera location and thus it concentrates its perturbation on these parts. As a result, RADA learns to generate efficient perturbations to the images, where these minimal perturbations are sufficient to perplex the network and can be used to improve the robustness to cross-weather variations.


\begin{figure}[h]
	\centering
	\includegraphics[width=\columnwidth]{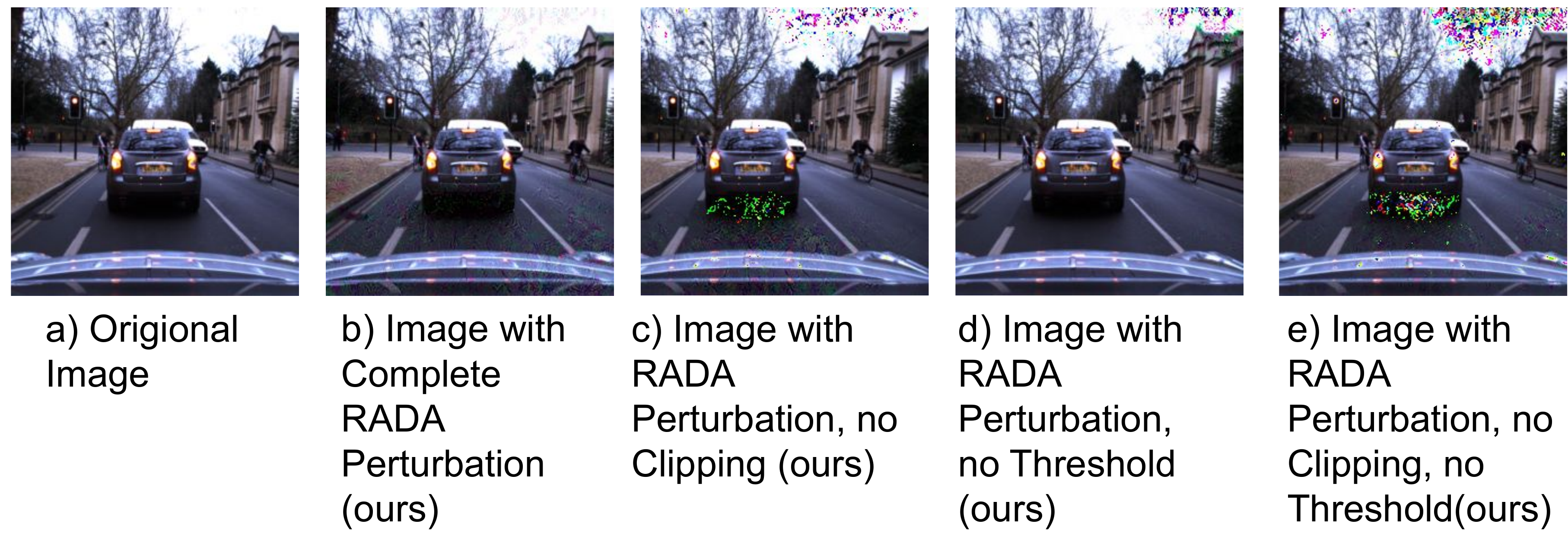}
	\caption{The perturbed images generated by RADA after turning off some building blocks. (a) Original image, (b) Complete version of RADA, (c) RADA without clipping mechanism, (d) RADA without threshold, and (e) RADA without threshold and clipping. It can be clearly seen that our proposed threshold and clipping mechanism can effectively avoid polluting crucial localization information or generating lots of confounding dots.}
	\label{fig:ablation_study}
\end{figure}



\begin{table}[h]

 \centering
 \caption{Ablation study of RADA on RobotCar (Basic AtLoc denotes the original AtLoc Model without RADA)}\label{tab:ablation}
 \resizebox{1\columnwidth}{!}{
 \begin{tabular}{l|l|l}
 \hline
 & Over-exposure & Rainy \\\hline
 Basic AtLoc&22.17, $17.72^{\circ}$&8.99, $2.15^{\circ}$\\\hline
 Complete RADA (ours)&\textbf{9.23m}, $\textbf{8.84}^{\circ}$&\textbf{6.87m}, $\textbf{1.90}^{\circ}$\\\hline
 No clipping RADA (ours)&12.39, $10.98^{\circ}$&10.96, $2.69^{\circ}$\\\hline
 No threshold RADA (ours)&11.99, $10.79^{\circ}$&8.82, $2.05^{\circ}$\\\hline
 no threshold, no clipping RADA (ours)&12.66, $12.75^{\circ}$&17.28, $2.95^{\circ}$\\\hline
 \end{tabular}}
\end{table}


\subsection{Ablation Study}
  We also conducted ablation study under different weather conditions using AtLoc model. In Table \ref{tab:ablation}, basic AtLoc is compared with a version trained with complete RADA, a clipping version of RADA trained without thresholding, a thresholding version trained without clipping, and a no-thresholding-no-clipping version of RADA. All models were trained in good condition and then tested in different `unseen' weather conditions. The rest settings are kept the same for fair comparison. The complete RADA achieved the best performance among all. This comparison indicates that all incomplete RADA versions are less accurate than the complete RADA, showing that the thresholding and clipping mechanism can produce effective constraints (see Fig. \ref{fig:ablation_study}). 


\subsection{Analysis on the Incremental Gain of RADA}

We further investigate what the relative gains from RADA are when supplied with a small proportion of data from the target domain e.g. a ratio of 80\%overcast to 20\% rainy. We tried different mixtures of dataset by incorporating 0 to 40\% of rainy day in the overcast training set. We then tested the model on two sets, one on the good (overcast) weather and another on the challenging rainy data. For each mixture of training data, we compute the mean error of basic AtLoc and the one trained with RADA. As shown in Table \ref{tab:partial_rain}, if we increase the percentage of rainy data from 0 to $40\%$ in the training set, RADA yields $68.3\%$ and $86.4\%$ smaller translation and rotation error respectively, especially when it is tested on rainy data. In summary, by adding a small portion of data from the target domain, it can significantly improve the accuracy of RADA.



\section{CONCLUSIONS}
We propose RADA, an adversarial data augmentation system for camera localization that can concentrate on perturbing the geometrically informative parts of the image. RADA is a simple plugin that can be used on virtually any deep-learning based camera localization network to generate adversarial training examples. RADA learns to generate minimal perturbations that can perplex the localization network and largely improve robustness. By using RADA, we demonstrated the possibility of using adversarial training to solve domain shift problem in camera localization. Through further experiments, we have also explored the trade-off between the quality of training dataset and the incremental benefit of RADA. For future work, an auto-encoder could be used to generate advanced adversarial perturbation in which the feature distribution is expected to be closer to the one generated from the real domain-shift problems. This might improve further the performance of RADA in a more challenging situation such as nighttime.

\medskip
{
\small
\bibliography{RADA}
\bibliographystyle{IEEEtran}
}
\end{document}